\documentclass[12pt,twoside]{article}   
\usepackage[super,numbers,sort]{natbib}
\usepackage{tabularx}
\usepackage{multirow}%
\usepackage{amsmath,amssymb,amsfonts}%
\usepackage{amsthm}%
\usepackage{mathrsfs}%
\usepackage[title]{appendix}%
\usepackage{xcolor}%
\usepackage{textcomp}%
\usepackage{manyfoot}%
\usepackage{booktabs}%
\usepackage{algorithm}%
\usepackage{algorithmicx}%
\usepackage{algpseudocode}%
\usepackage{listings}%
\usepackage{adjustbox}
\usepackage{multirow}  
\usepackage{adjustbox}
\usepackage{algorithm}  
\usepackage{algorithmicx}  
\usepackage{algpseudocode}  
\usepackage{amsmath} 
\usepackage{amsfonts}
\usepackage{booktabs}
\usepackage{multirow}
\usepackage{regexpatch}
\usepackage{amssymb}
\usepackage{mathrsfs}
\usepackage{makecell}
\usepackage[misc]{ifsym} 
\usepackage{placeins}
\usepackage{adjustbox}
\usepackage{caption}

\usepackage{hyperref}

\theoremstyle{thmstylethree}%
\usepackage{fancyhdr}		%Gives headers and footers defined below
\usepackage{graphicx}

\makeatletter \renewcommand\@biblabel[1]{$^{#1}$} \makeatother
\newcommand{\cen}[1]{\begin{center} #1 \end{center}}

\usepackage{hyperref}
\hypersetup{ colorlinks,
    citecolor=blue,
    filecolor=blue,
    linkcolor=blue,
    urlcolor=blue
}

\usepackage{xcolor}
\definecolor{gray}{rgb}{0.6,0.6,0.6}
\definecolor{red}{rgb}{0.85,0,0}
\definecolor{green}{rgb}{0,0.85,0}
\definecolor{blue}{rgb}{0,0,0.85}
\definecolor{beige}{rgb}{0.92,0.87,0.78}
\usepackage[all]{hypcap}    %causes link to figures to go to figure, not caption
\begin{document}

\cen{\sf {\Large {\bfseries nnSAM: Plug-and-play Segment Anything Model Improves nnUNet Performance } \\  
\vspace*{10mm}
Yunxiang Li\textsuperscript{a}, Bowen Jing\textsuperscript{a}, Zihan Li\textsuperscript{b}, Jing Wang\textsuperscript{a}, You Zhang\textsuperscript{a*}
} \\

\textsuperscript{a}Department of Radiation Oncology, UT Southwestern Medical Center, Dallas, 75390, TX, USA \\

\textsuperscript{b}Department of Bioengineering, University of Washington, Seattle, 98195, WA, USA \\

\textsuperscript{*}Author for correspondence: you.zhang@utsouthwestern.edu \\

}

\setcounter{page}{1}
\pagestyle{plain}

\begin{abstract}
\noindent {\bf Background:} The automatic segmentation of medical images has widespread applications in modern clinical workflows. The Segment Anything Model (SAM), a recent development of foundational models in computer vision, has become a universal tool for image segmentation without the need for specific domain training. However, SAM's reliance on prompts necessitates human-computer interaction during the inference process. Its performance on specific domains can also be limited without additional adaptation. In contrast, traditional models like nnUNet are designed to perform segmentation tasks automatically during inference and can work well for each specific domain, but they require extensive training on domain-specific datasets.\\ 
{\bf Purpose:} To leverage the advantages of both foundational and domain-specific models and achieve fully automated segmentation with limited training samples, we propose nnSAM, which combines the robust feature extraction capabilities of SAM with the automatic configuration abilities of nnUNet to enhance the accuracy and robustness of medical image segmentation on small datasets. \\
{\bf Methods:} We propose the nnSAM model for small sample medical image segmentation. We made optimizations for this goal via two main approaches: first, we integrated the feature extraction capabilities of SAM with the automatic configuration advantages of nnUNet, which enables robust feature extraction and domain-specific adaptation on small datasets. Second, during the training process, we designed a boundary shape supervision loss function based on level set functions and curvature calculations, enabling the model to learn anatomical shape priors from limited annotation data.\\
{\bf Results:} We conducted quantitative and qualitative assessments on the performance of our proposed method on four segmentation tasks: brain white matter, liver, lung, and heart segmentation. Our method achieved the best performance across all tasks. Specifically, in brain white matter segmentation using 20 training samples, nnSAM achieved the highest DICE score of 82.77 (± 10.12) \% and the lowest average surface distance (ASD) of 1.14 (± 1.03) mm, compared to nnUNet, which had a DICE score of 79.25 (± 17.24) \% and an ASD of 1.36 (± 1.63) mm. A sample size study shows the advantage of nnSAM becomes more prominent under fewer training samples.
{\bf Conclusions:} A comprehensive evaluation of multiple small-sample segmentation tasks demonstrates significant improvements in segmentation performance by nnSAM, highlighting the vast potential of small-sample learning. 
\end{abstract}

\newpage

\section{Introduction}\label{sec1}

Efficient and accurate segmentation of medical images is essential in modern clinical workflows for disease diagnosis and prognosis, treatment planning and monitoring, and treatment outcome follow-up\cite{liu2021review}. Traditionally, medical image segmentation is a very time-consuming and labor-intensive task. The advent of deep learning-based automatic segmentation techniques has significantly reduced the time and effort required from radiologists and radiation oncologists~\cite{sahiner2019deep}. Among the many deep learning architectures designed for biomedical image segmentation, UNet stands out for its ability to achieve good segmentation performance through a U-shaped design and skip connections\cite{ronneberger2015unet}. Based on the UNet backbone, a large number of studies further fine-tuned the network architectures with various modifications for different tasks~\cite{li2021gt,li2023lvit,zhou2019unet++,wang2024amsc,siddique2021u}. For example, Attention UNet introduces attention gates into UNet, allowing the model to automatically focus on target structures of varying shapes and sizes. However, these models are based on convolutional neural networks (CNNs), which are powerful in capturing local but not global information from images. To further improve the segmentation results, several transformer-based segmentation networks have been proposed. TransUNet integrates the advantages of CNNs and transformers in local and global feature extraction, which defines a new benchmark in medical image segmentation~\cite{chen2021TransUNet}. By utilizing the global contextual understanding of transformers and the precise localization capability of UNet, TransUNet can capture long-range dependencies while maintaining the segmentation accuracy of local structures. Another example is SwinUNet~\cite{cao2021swin}, which introduces a different transformer-driven approach to medical image segmentation, leveraging the U-shaped Encoder-Decoder architecture and skip-connections for enhanced local-global semantic feature learning. This model shows superior performance over both traditional convolution-based methods and mixed transformer-convolution techniques in multiple tasks. Many of the segmentation works, however, require substantial human effort in architecture modification and hyperparameter tuning to fit different applications or datasets. Acknowledging this challenge, the nnUNet framework~\cite{isensee2020nnUNet} was proposed. The nnUNet framework, a "no-new-UNet", takes a unique approach without proposing new network architectures. Instead, it refocuses efforts on methodological search and data processing steps to yield optimal performance. The nnUNet strategy demonstrates that with appropriate preprocessing and postprocessing combinations, even a basic network architecture can achieve state-of-the-art performance across a wide variety of medical segmentation tasks.

Historically, deep learning models for medical image segmentation, including nnUNet, were tailor-made for specific datasets or applications, making it challenging to generalize a single model’s effectiveness to various segmentation tasks. While the emergence of nnUNet signifies a transition to more flexible approaches for medical image segmentation, the quality of segmentation results still relies on ample training data on specific segmentation tasks. Acquiring large volumes of labeled medical images for each specific segmentation task is not only costly but also challenging in data-limited scenarios. For medical image segmentation tasks with a limited amount of training data, few-shot learning solutions, which allow new models to be trained based on a few samples, are important and more practical. The advent of the segment anything model (SAM)~\cite{kirillov2023segment,zhang2023faster}, a model that was trained on 11 million images and more than a billion segmentation masks (the SA-1B training dataset), has shown a great potential to achieve ‘few-shot’ and even ‘zero-shot’ learning across a diverse array of image categories. However, recent studies on the SAM model find its accuracy limited when applied directly to medical images without additional fine-tuning~\cite{mazurowski2023segment,ma2023segment}. In addition, the SAM model requires prompts as input in addition to the image itself (bounding box, points, etc.), which hinders its seamless integration in fully automated clinical workflows. This aspect, although a boon for human-machine interaction, may pose challenges in high-throughput medical scenarios that demand real-time or uninterrupted procedures. In order to realize fully automated medical image segmentation with SAM, AutoSAM~\cite{shaharabany2023autosam} is proposed, which directly learns prompts from input images and feeds the learned prompts into SAM for fully automated segmentation. However, AutoSAM needs to learn a new prompt encoder from the training dataset and is also susceptible to the scarcity of training data in few-shot scenarios.

Based on the above analysis, it is evident that SAM has the capability for zero-shot segmentation without additional training, but its inference process relies on human interaction, making it only semi-automatic. Its performance on specific medical datasets can also be sub-optimal. In contrast, fully automatic segmentation models like nnUNet typically start from scratch and require extensive, domain-specific training data to achieve good segmentation performance. Acknowledging their pros and cons, we attempt to implement a middle-ground solution that can achieve precise, fully automatic medical image segmentation with a small number of domain-specific training samples. Therefore, we propose nnSAM, a new plug-and-play solution designed to enhance the accuracy of medical image segmentation. nnSAM combines the robust feature extraction and generalization abilities of SAM with the data-centric automatic configuration abilities of nnUNet. By leveraging SAM's universal image encoder and seamlessly integrating it into the architecture of nnUNet, nnSAM produces a robust latent space representation, laying the foundation for improved segmentation precision. Additionally, allowing the model to learn more prior knowledge in situations of scarce training data helps to improve segmentation performance. To this end, we also designed a curvature loss based on the foundation of level sets~\cite{li2011level,wang2020deep,luo2021semi}, which helps the model to capture prior shape information from a small set of segmentation targets. 

The main contributions of this paper are as follows:
\begin{itemize}
    \item We introduce nnSAM, a novel fusion of the Segment Anything Model (SAM) and nnUNet. By combining the powerful feature extraction capabilities of SAM with the automatic configurable design of nnUNet, nnSAM enables improved segmentation quality under very limited training data.
    \item  We designed a curvature loss function based on level sets, enabling the model to learn the shape priors of segmentation targets for anatomically-reasonable inference.
    \item Our comprehensive evaluations show that nnSAM's performance surpasses existing state-of-the-art techniques, providing a potential new benchmark for medical image segmentation.
\end{itemize}

\begin{figure}[htb]
\includegraphics[width=\textwidth]{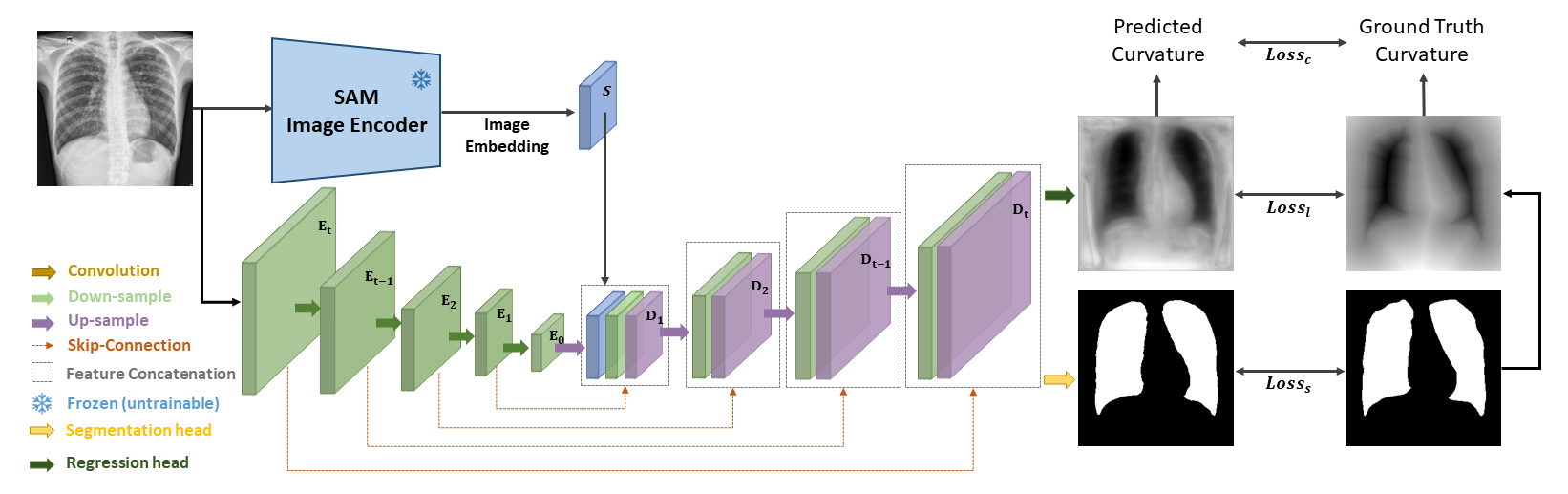}
\caption{The architecture of nnSAM. nnSAM integrates nnUNet’s encoder with the pre-trained SAM encoder. The correspondingly concatenated embeddings are input into nnUNet’s decoder, which has two output layers: a segmentation header, and a level set-based regression header. The segmentation header serves as the final output, while the regression header assists the model in capturing the shape priors during the training process.} 
\label{nnSAM}
\end{figure}

\section{Method}
\subsection{Architecture Overview}
The architecture of the proposed nnSAM framework is depicted in Figure 1. The model is designed to combine the strengths of nnUNet~\cite{isensee2020nnUNet} and SAM~\cite{kirillov2023segment}. Specifically, nnSAM consists of two parallel encoders: the nnUNet encoder and the SAM encoder. The SAM encoder is a pre-trained Vision Transformer (ViT)~\cite{dosovitskiy2020image}. The embeddings from both encoders are concatenated and subsequently fed into nnUNet’s decoder. The decoder has two output layers, one is a segmentation head, and another is a level set-based regression head. The segmentation head is trained with cross-entropy loss and DICE loss, while the regression head is trained with MSE loss and a proposed curvature loss. The SAM encoder is used as a plug-and-play plugin, with its parameters frozen during training. Accordingly, only the weights of the nnUNet encoder and decoder are updated during the training process.

\subsection{Auto-configured nnUNet Architecture}
Integrating nnUNet into the nnSAM framework allows automated network architecture and hyperparameter configuration, making it highly adaptable to the unique and specific features of each medical imaging dataset. This adaptive capability starts from a self-configuration process that automatically adjusts the nnUNet encoder’s architecture, including parameters such as the layer count and the convolutional kernel size, to suit training dataset characteristics including the dimensions of the input images, the number of channels, and the number of classes involved in the segmentation task. Additionally, nnUNet uses an automated preprocessing pipeline, which includes normalizing the input data and applying data augmentation techniques such as rotations, scaling, and elastic deformations. These preprocessing and augmentation steps are crucial for improving the robustness and accuracy of the model. Beyond these, nnUNet can automatically adjust optimizer settings based on the dataset’s inherent attributes. For example, nnUNet can automatically optimize key hyperparameters including the learning rate and the batch size. The comprehensive suite of auto-configurable features allows the nnUNet and correspondingly the nnSAM architecture to optimize its setup for each specific medical imaging task, enhancing both its efficiency and accuracy. Since the number of layers of the nnSAM is determined by the specific dataset, in Figure ~\ref{nnSAM}, we symbolize the number of encoder layers as $E_t$ to $E_0$ and the number of decoder layers as $D_1$ to $D_t$.

\subsection{SAM Encoder}

The SAM encoder is a Vision Transformer model pre-trained on the extensive SA-1B segmentation dataset. Trained with this extremely large dataset, the SAM encoder excels at domain-agnostic feature extraction for segmentation tasks. However, its segmentation ability is highly prompt-dependent, making it unable to self-identify the segmentation target and the underlying semantics. Therefore, nnSAM only uses the SAM encoder to incorporate its feature extraction strengths, while leaving the dataset-specific task (identifying the region of interest for segmentation) to nnUNet. For an input image $x \in \mathbb{R}^{H \times W \times N} $, where $ H \times W $ are the spatial dimensions and $N$ is the number of channels, the SAM encoder needs the input $ H \times W $ to be of size 1024$\times$1024. To meet this requirement, we resize the corresponding image dimension to 1024$\times$1024 using linear interpolation after the pre-processing of nnUNet. The SAM encoder produces an image embedding $ S $ with dimensions 64$\times$64. We subsequently resize this embedding $S$ to match the dimensions of nnSAM’s decoder layer $ D_1 $ for concatenation (Figure 1). 
To balance the inference speed of nnSAM with the segmentation accuracy, we use MobileSAM~\cite{zhang2023faster,wu2022tinyvit}, a lightweight SAM version that is less than 1/60 in size of the original SAM, but with comparable performance. MobileSAM is obtained by distillation from the original SAM, through which the knowledge from the original image encoder is transferred into the lightweight counterpart.

\subsection{Multi-head Decoder}
After obtaining the embeddings from the nnUNet encoder and the SAM encoder, they are input together into our multi-head decoder, which consists of a pixel-probability-based segmentation head and a regression head for level set representation. The pixel-probability-based segmentation head is a common approach in segmentation networks, generating the final segmentation mask. It is trained using the segmentation loss $Loss_s$, which is a combination of the DICE loss and the cross-entropy (CE) loss, defined as follows:

\begin{equation}
L_{\text{DICE}} = 1 - \frac{ 2\sum_{a=1}^{H}\sum_{b=1}^{W} \sum_{j=1}^{C} p_{j}(a,b) y_{j}(a,b)}{\sum_{a=1}^{H}\sum_{b=1}^{W} \sum_{j=1}^{C} (p_{j}(a,b) + y_{j}(a,b))}
\end{equation}

\begin{equation}
L_{\text{CE}} = -\frac{1}{HWC}\sum_{a=1}^{H}\sum_{b=1}^{W} \sum_{j=1}^{C}   y_{j}(a,b) \log (p_{j}(a,b))
\end{equation}

\begin{equation}
Loss_{s} = L_{\text{DICE}} + L_{\text{CE}}
\end{equation}

where $H$ and $W$ indicate height and width; $a, b$ denote the coordinate positions; $C$ represents the number of segmentation classes; $p_{j}(a,b)$ is the inferred probability that pixel $(a,b)$ belongs to class $j$; and $y_{j}(a,b)$ denotes the ground truth, which indicates whether pixel $(a,b)$ belongs to class $j$. For the regression header, a level set function related to the boundary contour is used to provide curvature supervision~\cite{chen2020learning} and assist in learning the shape priors under limited sample conditions. The level set function is defined as follows:

\begin{equation}
\phi(a, b) = \begin{cases} 
-d(a, b) & \text{if } (a, b) \text{ is inside the object} \\
0 & \text{if } (a, b) \text{ is on the boundary of the object} \\
d(a, b) & \text{if } (a, b) \text{ is outside the object}
\end{cases}
\end{equation}

where $\phi(a, b)$ is the level set function and $d(a, b)$ denotes the minimum distance from the point $(a, b)$ to the nearest object boundary. With the above equation, we can convert the segmentation ground truth $y_{j}(a,b)$ into the level set ground truth $\phi_{j}(a, b)$. The MSE loss between the ground truth $\phi_{j}(a, b)$ and the predicted level set $\phi_{j}'(a, b)$ is calculated as:

\begin{equation}
\text{Loss}_l = \text{MSE}(\phi(a, b), \phi'(a, b)) = \frac{1}{HWC}\sum_{a=1}^{H}\sum_{b=1}^{W} \sum_{j=1}^{C}  (\phi_j(a, b) - \phi_j'(a, b))^2
\end{equation}

The above loss measures the distance map discrepancy of the level set function, which is directly correlated with the segmentation. To further capture the curvature of the ground-truth segmentation to learn the shape priors, we turn the boundary of the ground-truth level set into the region with the largest gradient by a Sigmoid function.

\begin{equation}
\hat{\phi}(a,b) = Sigmoid(-1000 \times \phi(a,b))
\end{equation}

\begin{equation}
\hat{\phi}_a = \frac{\partial \hat{\phi}(a,b)}{\partial a}, \quad \hat{\phi}_b = \frac{\partial \hat{\phi}(a,b)}{\partial b}
\end{equation}

\begin{equation}
\hat{\phi}_{aa} = \frac{\partial^2 \hat{\phi}(a,b)}{\partial a^2}, \\
\hat{\phi}_{bb} = \frac{\partial^2 \hat{\phi}(a,b)}{\partial b^2}, \\
\hat{\phi}_{ab} = \frac{\partial^2 \hat{\phi}(a,b)}{\partial a \partial b}
\end{equation}

where \( \hat{\phi}_a \) and \( \hat{\phi}_b \) are the first derivatives of \( \hat{\phi}(a,b) \) with respect to a and b. \( \hat{\phi}_{aa} \), \( \hat{\phi}_{bb} \), and \( \hat{\phi}_{ab} \) are the second derivatives along a, b, and mixed a-b, respectively.
Following this, we can calculate the curvature of the ground-truth segmentation by:

\begin{equation}
K_{\hat{\phi}} = \frac{\left| (1+\hat{\phi}_a^2) \hat{\phi}_{bb} + (1+\hat{\phi}_b^2) \hat{\phi}_{aa} - 2 \hat{\phi}_a \hat{\phi}_b \hat{\phi}_{ab} \right|}{2(1+ \hat{\phi}_a^2 + \hat{\phi}_b^2)^{1.5}} \\
\end{equation}

Similarly, we can obtain the curvature $K_{\hat{\phi'}}$ of the predicted level sets, and can compute their differences in boundary curvature as $Loss_{c}$:

\begin{equation}
Loss_{c} = | K_{\hat{\phi}} - K_{\hat{\phi'}}|
\end{equation}

The final loss function for model training is:
\begin{equation}
Loss = \lambda_1 \times Loss_{s} + \lambda_2 \times Loss_{l} + \lambda_3 \times Loss_{c} 
\end{equation}

where $\lambda_1$, $\lambda_2$, and $\lambda_3$ are the weights that can be adjusted according to the importance of each component.

\section{Experimental Setting}
We evaluated nnSAM using four tasks, including substructure segmentation of the heart on CT scans~\cite{zhuang2019evaluation}, white matter segmentation on MR scans~\cite{crimi2019brainlesion}, Chest X-ray segmentation~\cite{jaeger2013automatic}, and liver segmentation on CT scans~\footnote{ https://doi.org/10.7303/syn3193805}. All data were processed into 2D slices, with each image resized to 256×256. To assess the performance of nnSAM under limited sample training, we divided all tasks into 20 training samples and 100 validation samples. There were 200 test samples for substructure segmentation of the heart on CT, and 400 test samples each for white matter segmentation on MR, Chest X-ray segmentation, and liver segmentation on CT. To further evaluate nnSAM's performance under varying levels of training data scarcity, we trained different versions of nnSAM using varying sizes of training sample subsets for the MR white matter segmentation, ranging from 5 to 20 samples. This allowed us to study how nnSAM's performance varies with the size of available labeled data, simulating real clinical environments where labeled data may be difficult to obtain.

In addition to nnSAM, we also evaluated UNet~\cite{ronneberger2015unet}, Attention UNet~\cite{oktay2022attention}, SwinUNet~\cite{cao2021swin}, TransUNet~\cite{chen2021TransUNet}, AutoSAM~\cite{shaharabany2023autosam}, and the original nnUNet~\cite{isensee2020nnUNet}, for comparison.  For UNet, Attention UNet, SwinUNet, TransUNet, and the original nnUNet, we used the available public codes. While for AutoSAM, since there is no official open-source code, we have reproduced it based on the article descriptions. For the evaluation metric, we used Average Symmetric Surface Distance (ASD) and the DICE Similarity Coefficient (DICE)~\cite{li2021agmb}. The ASD is a metric that quantifies the average distance between the surfaces of two segmented objects. DICE evaluates the similarity between two segmented objects, considering the volume overlap between the two objects. Considering that our model consists of a regression head and a segmentation head, where the regression head's role is to help the model learn more anatomical shape priors during training, we set the weight of the regression head's loss $\lambda$ to be smaller. The $\lambda_1$, $\lambda_2$, and $\lambda_3$ used during training were 1, 0.1, and 0.0001, respectively.

\begin{figure}[ht!]
\centering
\includegraphics[width=\textwidth]{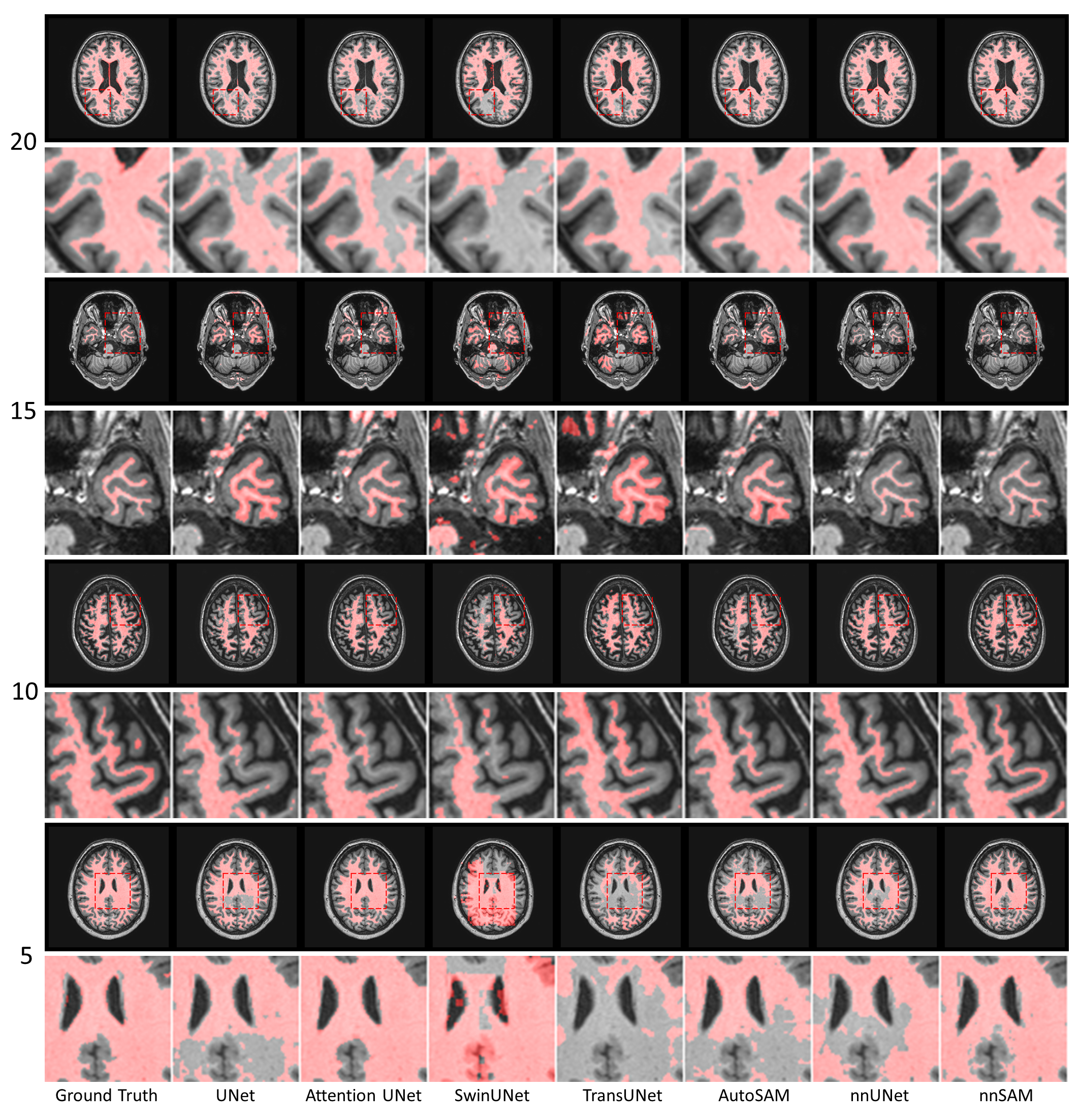}
\caption{Segmentation visualization results for different methods on MR brain white matter segmentation, with the numbers on the left representing different training sample sizes. For the displayed images of each training sample size, a full segmentation (upper row) and a zoomed-in segmentation (lower row) are shown. } 
\label{fig:res1}
\end{figure}

\begin{table}[t!]
  \small
  \caption{DICE and ASD of different MR brain white matter segmentation methods under various training sample sizes. }
  \centering
  \begin{adjustbox}{center}
  \begin{tabular}{c|c|c|c|c|c}
  \hline
  
  Method & Metrics & \textbf{5} & \textbf{10} & \textbf{15} & \textbf{20} \\ \hline
  
  \multirow{2}{*}{\textbf{UNet}} 
  &DICE (\%) & $68.14 \pm 22.91$ & $73.14 \pm 11.95 $ & $76.24 \pm 14.12$ & $76.85 \pm 12.29$ \\
  &ASD (mm)& $3.71 \pm 3.54$ & $2.74 \pm 2.82$ & $2.24 \pm 2.59$ & $2.35 \pm 2.98$ \\
  \hline
  
  \multirow{2}{*}{\textbf{Attention UNet}} 
  &DICE (\%) & $66.26 \pm 23.28$ & $73.52 \pm 16.10$ & $77.49 \pm 11.62$ & $76.73 \pm 13.10$ \\
  &ASD (mm)& $4.02 \pm 3.89$ & $2.35 \pm 2.44$ & $1.85 \pm 2.75$ & $2.47 \pm 2.68$ \\
  \hline
  
  \multirow{2}{*}{\textbf{SwinUNet}} 
  & DICE (\%) & $41.02 \pm 21.47$ & $58.69 \pm 14.01$ & $70.33 \pm 13.06$ & $73.09 \pm 12.32$ \\
  & ASD (mm)& $9.05 \pm 6.56$ & $3.71 \pm 2.80$ & $2.57 \pm 2.13$ & $2.12 \pm 1.90$ \\ \hline
  
  \multirow{2}{*}{\textbf{TransUNet}} 
  & DICE (\%) & $65.18 \pm 18.46$ & $72.43 \pm 13.01$ & $74.30 \pm 12.14$ & $77.66 \pm 11.86$ \\
  & ASD (mm)& $3.42 \pm 2.94$ & $2.26 \pm 2.32$ & $1.93 \pm 1.77$ & $1.68 \pm 1.94$ \\ \hline
  
  \multirow{2}{*}{\textbf{AutoSAM}} 
  & DICE (\%) & $68.28 \pm 19.95$ & $74.11 \pm 12.39$ & $77.47 \pm 10.72$ & $77.44 \pm 14.69$ \\
  & ASD (mm)& $3.67 \pm 3.61$ & $2.08 \pm 2.70$ & $1.90 \pm 1.88$ & $1.69 \pm 1.55$ \\ \hline
  
  \multirow{2}{*}{\textbf{nnUNet}} 
  & DICE (\%) & $68.25 \pm 25.66$ & $74.74 \pm 21.53$ & $77.83 \pm 17.32$ & $79.25 \pm 17.24$ \\
  & ASD (mm)& $2.14 \pm 3.21$ & $1.77 \pm 1.61$ & $1.32 \pm 1.07$ & $1.36 \pm 1.63$ \\ \hline
  
  \multirow{2}{*}{\textbf{nnSAM}} 
  & DICE (\%) & \textbf{74.55 $\pm$ 19.93} & \textbf{78.50 $\pm$ 14.08} & \textbf{80.82 $\pm$ 13.44} & \textbf{82.77 $\pm$ 10.12} \\
  & ASD (mm)& \textbf{2.06 $\pm$ 2.88} & \textbf{1.56 $\pm$ 1.98} & \textbf{1.23 $\pm$ 1.01} & \textbf{1.14 $\pm$ 1.03} \\ \hline
  
  \end{tabular}
  \end{adjustbox}
  \label{tab:brain}
  \end{table}

\vspace*{6mm}
\section{Results}
\subsection{MR White Matter Segmentation}
Table~\ref{tab:brain} displays the performance of the model for MR brain white matter segmentation across different numbers of training data samples (from 5 to 20). Across all sample sizes, the proposed nnSAM outperformed all other segmentation methods in both DICE and ASD metrics. When trained with 20 annotated images, nnSAM achieved an average DICE score of 82.77\% and an average ASD of 1.14 mm. nnUNet, recognized as one of the best segmentation models, also significantly outperformed the other methods, though inferior to nnSAM. Other methods including SwinUNet, TransUNet, and AutoSAM showed much lower accuracy, with DICE scores below 80\%. SwinUNet performed the worst, which was expected, as its main structure composed of transformers which generally require a large amount of training data to work properly \cite{dosovitskiy2020image}, rendering it most affected by insufficient training samples. TransUNet performed relatively better as it incorporates a transformer block only at the bottom of the model while the main architecture remains CNN-based. Moreover, nnSAM's advantage over other methods becomes more pronounced with the reduction in the number of training samples. Notably, when trained with only 5 annotated images, nnSAM scored about 6.3\% higher in DICE than the second-ranked nnUNet and significantly higher than other methods. Consistent with the quantitative results, Figure \ref{fig:res1} shows that nnUNet and nnSAM perform similarly with 20 samples, but as the number of samples decreases, the performance gap between nnUNet and nnSAM becomes wider. Another point of note is that AutoSAM performs moderately with 20 samples but ranks second with a DICE of 68.28 when the sample count is reduced to 5, proving that the SAM encoder can provide significant domain-agnostic assistance under small sample conditions.
Overall, nnSAM showed higher segmentation accuracy compared with other methods, especially under conditions of limited training data.

\begin{figure}[ht!]
\centering
\includegraphics[width=\textwidth]{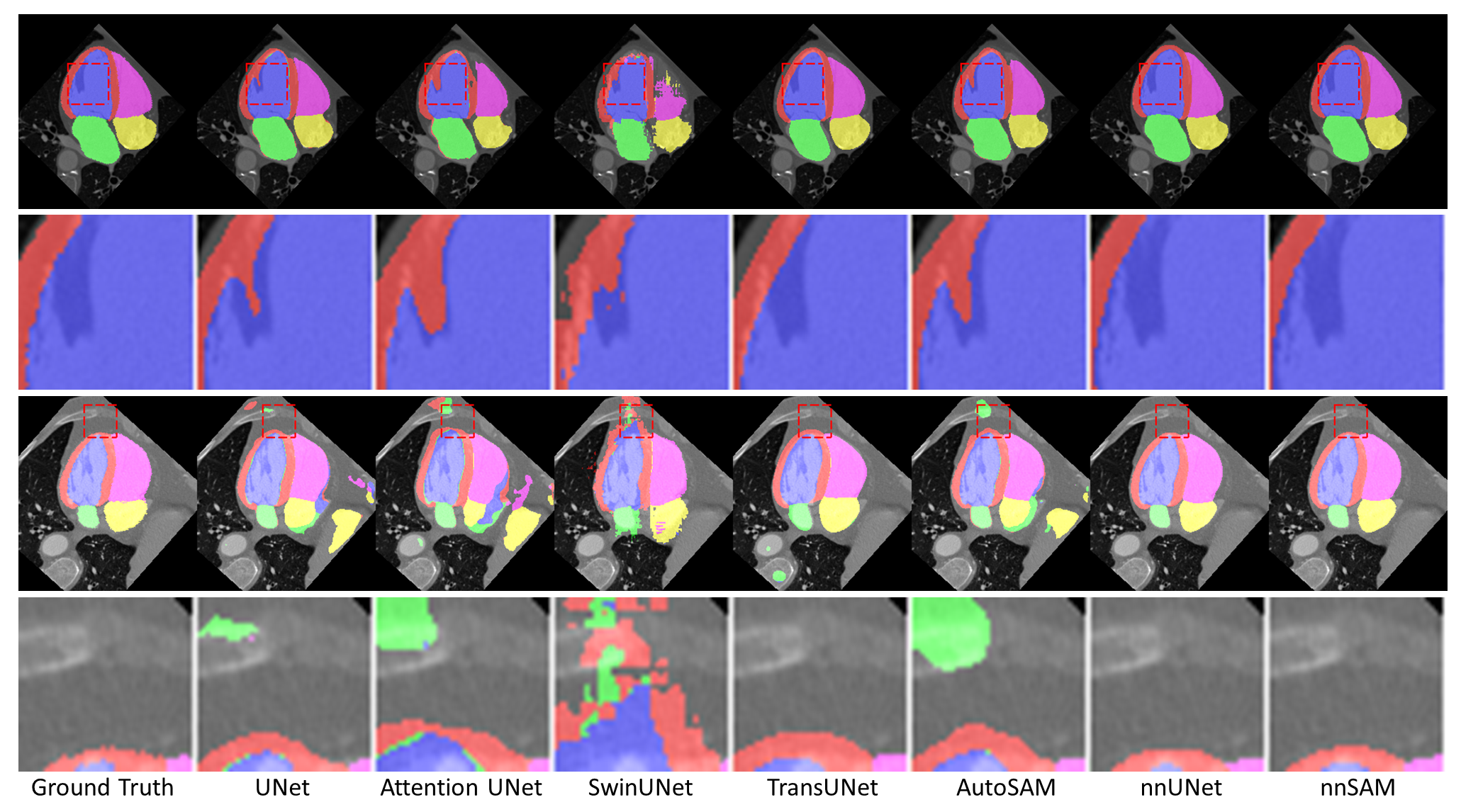}
\caption{Segmentation visualization results for different methods on CT heart substructure segmentation.} 
\label{fig:res2}
\end{figure}

\begin{table}[htb!]
\centering
\setlength{\tabcolsep}{12pt}
\caption{DICE and ASD of different methods on CT heart substructure segmentation.}
\begin{tabular}{c|c|c}
\hline
\textbf{Method} & \textbf{DICE (\%)} & \textbf{ASD (mm)} \\
\hline
UNet & 89.74 $\pm$ 2.96 & 5.96 $\pm$ 1.67 \\
Attention UNet & 87.99 $\pm$ 4.25  & 5.85 $\pm$ 1.78  \\
SwinUNet & 82.2 $\pm$ 6.39 & 4.23 $\pm$ 1.53 \\
TransUNet &  89.43  $\pm$ 2.68 & 2.41 $\pm$ 1.15 \\
AutoSAM & 90.29 $\pm$ 3.1 & 4.36 $\pm$ 1.41 \\
nnUNet & 93.76 $\pm$ 2.95 & 1.48 $\pm$ 0.65 \\
\textbf{nnSAM} & \textbf{94.19} $\pm$ \textbf{1.51} & \textbf{1.36} $\pm$ \textbf{0.42} \\

\hline
\end{tabular}
\label{tab:heart}
\end{table}

\subsection{CT Heart Substructure Segmentation}
Table \ref{tab:heart} and Figure \ref{fig:res2} show the performance in DICE and ASD for CT heart substructure segmentation. In the first (top) visualization example of Figure \ref{fig:res2}, many methods including UNet, Attention UNet, SwinUNet, and AutoSAM incorrectly segmented the shadowed area. while nnSAM managed to segment it accurately. In the second (bottom) visualization example of Figure \ref{fig:res2}, SwinUNet remains the worst-performing model, consistent with its performance in the MR brain white matter segmentation task, highlighting its inadequacy under small sample conditions. Overall, nnSAM still achieved the best results, and nnUNet remained the second-ranked model, closest to nnSAM. AutoSAM has the third-best DICE score, but performs poorly in ASD, showing its domain-agnostic potential but the lack of domain-specific fine-tuning.

\begin{figure}[t!]
\centering
\includegraphics[width=\textwidth]{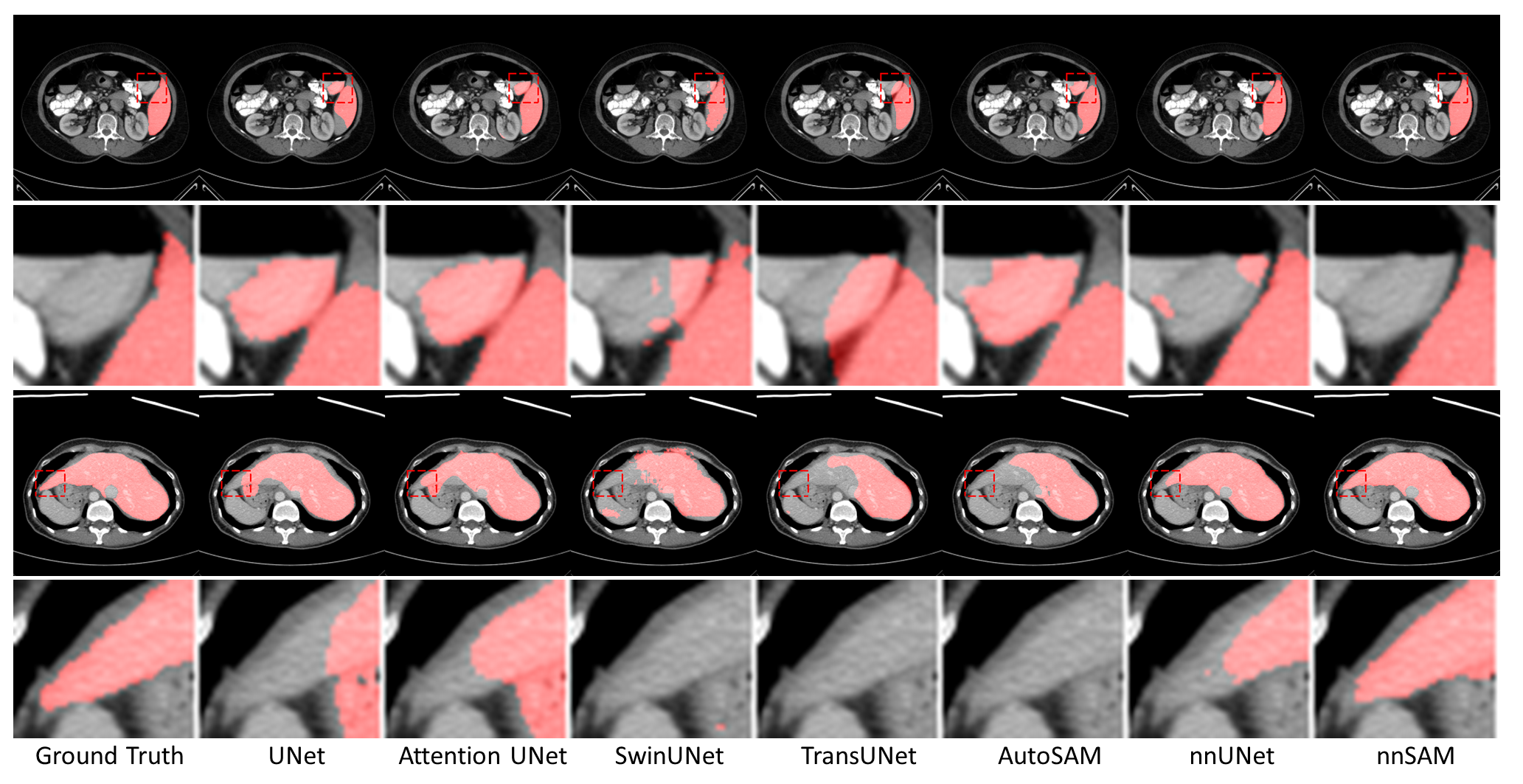}
\caption{Segmentation visualization results for different methods on CT liver segmentation.} 
\label{fig:res3}
\end{figure}

\begin{table}[t!]
\centering
\setlength{\tabcolsep}{12pt}
\caption{DICE and ASD of different methods on CT liver segmentation.}
\begin{tabular}{c|c|c}
\hline
\textbf{Method} & \textbf{DICE (\%)} & \textbf{ASD (mm)} \\
\hline
UNet & 82.4 $\pm$ 18.1 & 6.4 $\pm$ 6.42 \\
Attention UNet & 82.28 $\pm$ 17.22 & 7.18 $\pm$ 6.87 \\
SwinUNet & 74.47 $\pm$ 22.13 & 8.0 $\pm$ 6.31 \\
TransUNet & 74.97 $\pm$ 20.14 & 8.5 $\pm$ 5.93 \\
AutoSAM & 82.94 $\pm$ 16.56 & 5.98 $\pm$ 4.82 \\
nnUNet & 83.69 $\pm$ 26.32 & 6.7 $\pm$ 15.66 \\
\textbf{nnSAM} & \textbf{85.24} $\pm$ \textbf{23.74} & \textbf{6.18} $\pm$ \textbf{16.02} \\
\hline
\end{tabular}
\label{tab:liver}
\end{table}

\subsection{CT Liver Segmentation}
The comparisons of liver segmentation on CT scans in Figure \ref{fig:res3} show that nnSAM, empowerd by the regression head, can effectively address false positives and false negatives by learning shape priors. For instance, in the first (top) example in Figure \ref{fig:res3}, most models except nnSAM incorrectly predicted the adjacent structure as liver. However, nnSAM, having learned the shape prior, was able to infer a distinct boundary matching the ground truth. Similarly, in the second (bottom) example in Figure \ref{fig:res3}, nnSAM accurately predicted the tip region of the liver, while other models missed this part. By learning the liver's shape information through the regression head, nnSAM can generate anatomically reasonable results during segmentation tasks. Consistent with the performance depicted in the figures, nnSAM performed the best in Table \ref{tab:liver}, achieving the highest scores in both DICE and ASD.

\subsection{X-ray Chest Segmentation}
\begin{figure}[htb!]
\centering
\includegraphics[width=\textwidth]{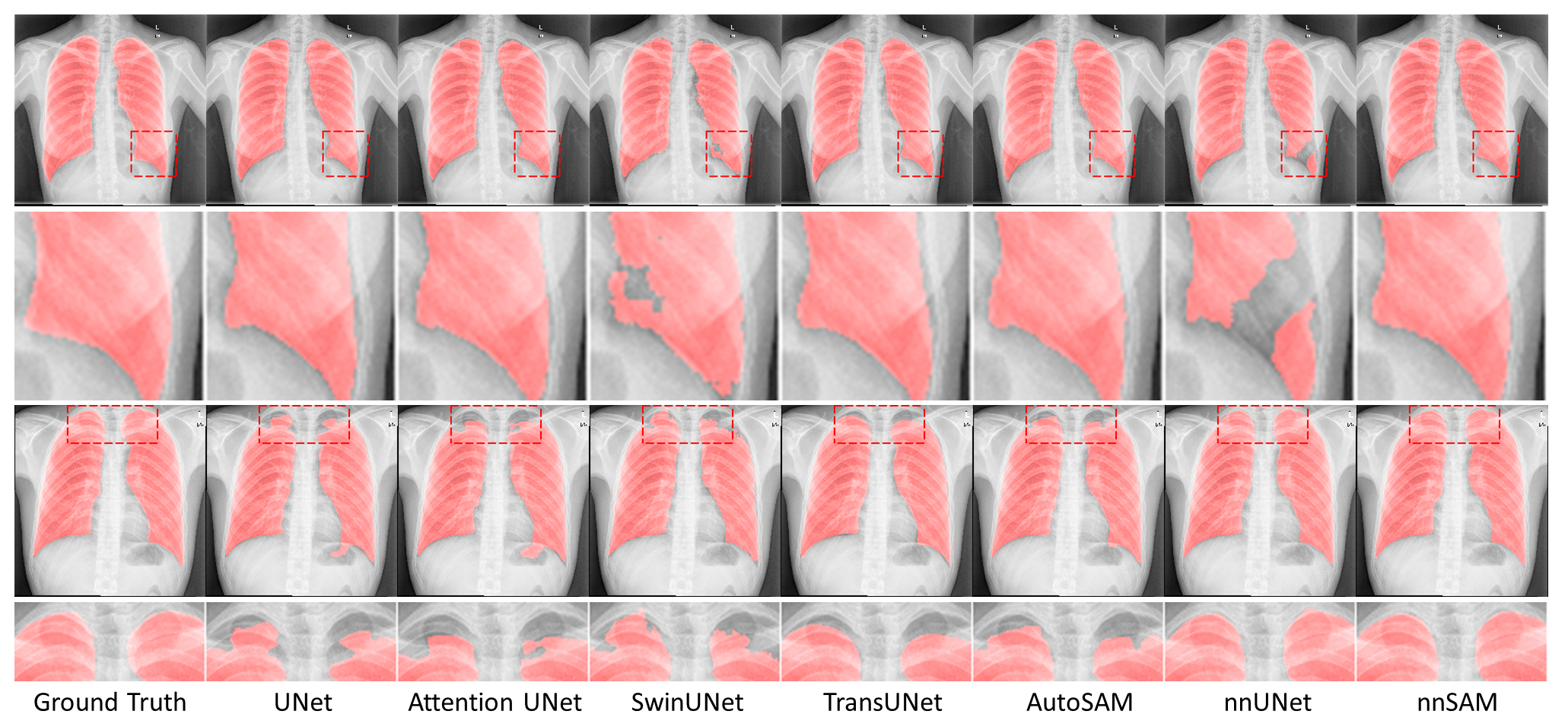}
\caption{Segmentation visualization results for different methods on chest X-ray segmentation.} 
\label{fig:res4}
\end{figure}

Given that the shape of most lungs follows certain patterns, chest X-ray segmentation is a task that benefits significantly from strong shape priors. In the first example (top) from Figure \ref{fig:res4}, nnUNet's results show a missing area, whereas nnSAM correctly captured it. Regarding the second example (bottom) from Figure 5, the top of the lungs typically presents a smooth, curved boundary, yet most comparison models, lacking a regression head to learn shape priors, yielded incomplete boundaries. Table \ref{tab:lung} also shows that nnUNet remains the second-ranked model, just behind nnSAM, which attests to the usefulness of nnUNet's framework that automatically configures the network based on the dataset in medical image segmentation tasks.

Overall, across four tasks, nnSAM demonstrates excellent accuracy with only a small number of training samples when segmenting challenging targets. The improvement over other state-of-the-art techniques is owning to the robust feature extraction of the SAM encoder and the adaptive capability of the nnUNet auto-configurable framework, as well as the regression head we designed to help the model to capture the shape prior from limited samples for additional regularization.

\begin{table}[htb!]
\centering
\setlength{\tabcolsep}{12pt}
\caption{DICE and ASD of different methods on chest X-ray segmentation.}
\begin{tabular}{c|c|c}
\hline
\textbf{Method} & \textbf{DICE (\%)} & \textbf{ASD (mm)} \\
\hline
UNet & 88.28 $\pm$ 3.46 & 3.61 $\pm$ 1.56 \\
Attention UNet & 87.73 $\pm$ 4.2 & 4.58 $\pm$ 2.42 \\
SwinUNet & 79.9 $\pm$ 6.06 & 3.83 $\pm$ 1.2 \\
TransUNet & 87.5 $\pm$ 2.8 & 2.89 $\pm$ 1.3 \\
AutoSAM & 88.06 $\pm$ 3.26 & 4.62 $\pm$ 1.52 \\
nnUNet & 93.01 $\pm$ 2.41 & 1.63 $\pm$ 0.57 \\
\textbf{nnSAM} & \textbf{93.63} $\pm$ \textbf{1.49} & \textbf{1.47} $\pm$ \textbf{0.42} \\
\hline
\end{tabular}
\label{tab:lung}
\end{table}

\begin{figure}[htb!]
  \centering
  \includegraphics[width=.5\textwidth]{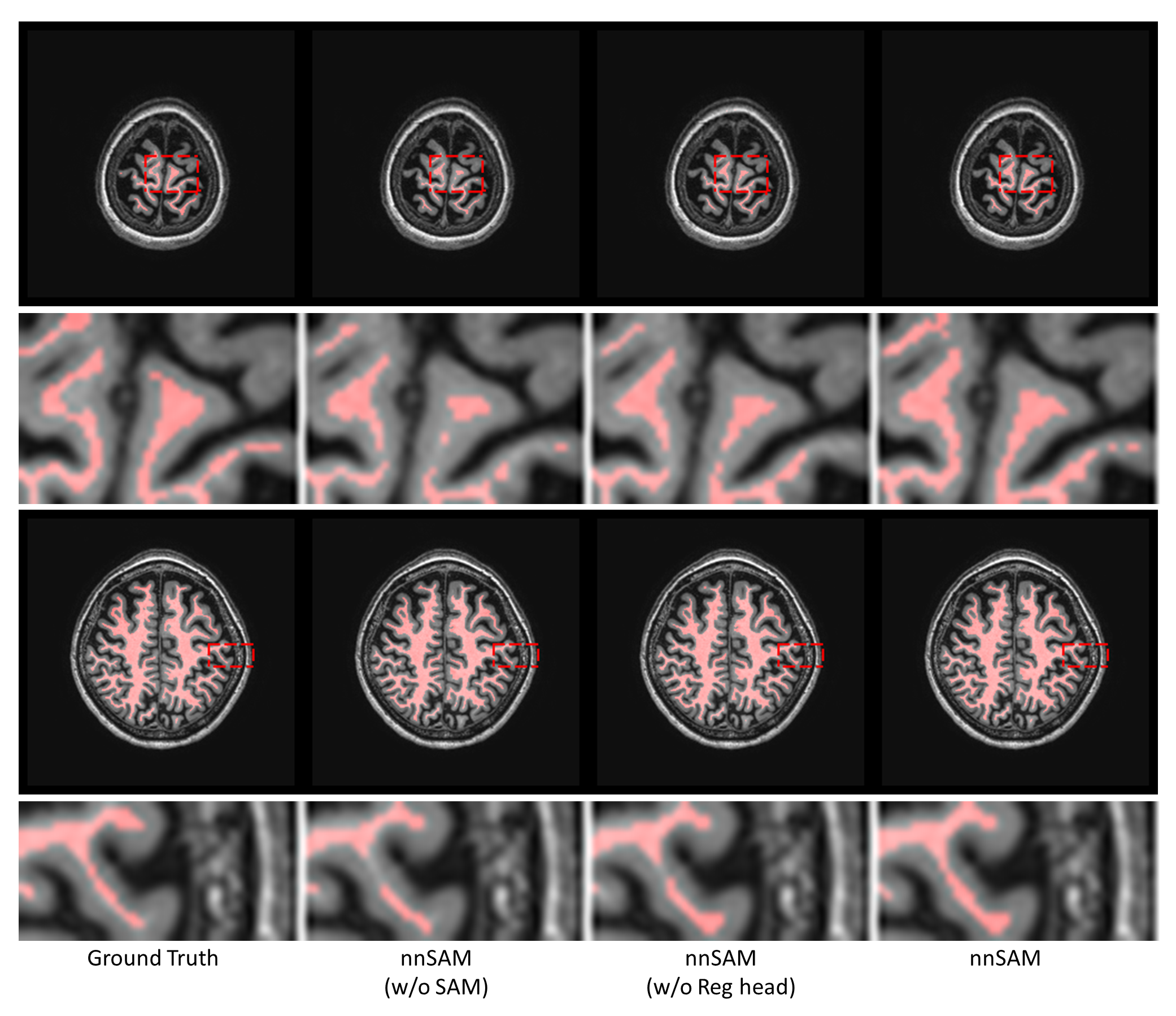}
  \caption{Segmentation visualization results after removing different components of nnSAM. (w/o: without)} 
  \label{fig:res5}
\end{figure}

\begin{table}[htb!]
\centering
\setlength{\tabcolsep}{12pt}
\caption{Ablation study on MR brain white matter segmentation. (w/o: without)}
\begin{tabular}{c|c|c}
\hline
\textbf{Method} & \textbf{DICE (\%)} & \textbf{ASD (mm)} \\
\hline
nnSAM (w/o SAM) & 81.1 $\pm$ 12.6 & 1.16 $\pm$ 1.24 \\
nnSAM (w/o Reg head) & 80.63 $\pm$ 13.46 & 1.34 $\pm$ 1.49 \\
\textbf{nnSAM}           & \textbf{82.77} $\pm$ \textbf{10.12} & \textbf{1.14} $\pm$ \textbf{1.03} \\
\hline
\end{tabular}
\label{tab:ablation}
\end{table}

\subsection{Ablation Study}
Through the experiments above, we have demonstrated the strong performance of nnSAM on small sample size data. To verify the contribution of various components to the final performance, we conducted ablation studies by separately removing the SAM encoder and the regression head to demonstrate their effectiveness. As seen from Table \ref{tab:ablation} and Figure \ref{fig:res5}, the model's performance declines when either the SAM encoder or the regression head is removed. Notably, after removing the regression head, there is a significant increase in ASD. The shape prior learned through the regression head helps to smooth the boundaries of the segmentation results, which significantly impacts ASD which is calculated based on the boundary distance.

\section{Discussion}

The study results demonstrate the superior performance of nnSAM in medical image segmentation, particularly in few-shot learning scenarios where labeled training data is limited. By integrating the pretrained SAM encoder into the nnUNet framework, nnSAM can leverage SAM's powerful feature extraction capabilities while benefiting from nnUNet's adaptive architecture configuration and hyperparameter optimization. The evaluations conducted across four tasks highlight several key advantages of nnSAM. Firstly, nnSAM consistently maintains the highest accuracy across various tasks, surpassing advanced models such as nnUNet, SwinUNet, and TransUNet. Secondly, in the segmentation of MR brain white matter, we tested different training set sizes (from 5 to 20 samples) and found that nnSAM's advantages over other models become more pronounced with smaller training data sets. This demonstrates that nnSAM can achieve accurate segmentation results from very few samples, making it highly valuable in medical applications where acquiring labeled data is both difficult and expensive. Compared to nnSAM, AutoSAM uses a custom encoder to replace the prompting encoder, enabling it to autonomously generate and provide cues to SAM. However, AutoSAM is not optimized for semantic segmentation of medical images like nnUNet, nor does it possess nnUNet’s powerful preprocessing and auto-configuration features, which affected its accuracy. Since the advent of nnUNet, it has become the state-of-the-art technology for most medical image segmentation tasks, representing a top-tier end-to-end model for traditional task-specific semantic segmentation. On the other hand, SAM is a prompt-based segmentation framework and also a representative model with strong generalizability. Our study has proved that combining the best models of two different segmentation frameworks can effectively further improve the accuracy of medical image segmentation. Additionally, for small-sample segmentation, we believe that enabling the model to learn additional shape priors will help produce more anatomically reasonable segmentation results, which was demonstrated in our results, with nnSAM preserving the shapes and smooth boundaries of structures including the liver and the lung. By the multi-head design, nnSAM learns the shape prior simultaneously during the end-to-end training, without needing to train a separate network to encode the shape information.

\section{Limitation}
Our current study has some limitations that should be addressed in future work. First, our method of using a regression head to learn shape priors may only be effective for the segmentation of organs or tissues with regular shapes and may not be suitable for the highly variable shapes of tumor segmentation. Secondly, the current nnSAM framework still requires a limited number of training data and labels, and future work needs to explore the possibilities of achieving end-to-end segmentation with only one sample ("one-shot" learning) or without any labels at all ("zero-shot" learning). Additionally, we train and test the models using two-dimensional slices. Expanding to three-dimensional, volume-based segmentation could further enhance segmentation accuracy by utilizing the correlation along the other direction, but there are currently technical difficulties in integrating three-dimensional SAM embeddings and three-dimensional nnUNet embeddings.

\section{Conclusion}
We introduced nnSAM, a novel few-shot learning solution for medical image segmentation that merges the advantages of the Segment Anything Model (SAM) and nnUNet, and incorporates a levelset-based regression head during training to help the model to capture shape priors. We conducted extensive evaluations on four segmentation tasks, establishing a potential new benchmark for medical image segmentation, particularly in scenarios with scarce training data. 

\section{Acknowledgement}
We thank Xiang Feng and Yongbo He for their helpful discussions at the early stages of the project. 

\section{Funding support}
The study was supported by funding from the National Institutes of Health (R01 CA240808, R01 CA258987, and R01 CA280135).

\section*{References}
\vspace{-10 mm}
\bibliography{example}   

\begin{thebibliography}{10}

\bibitem{liu2021review}
X.~Liu, L.~Song, S.~Liu, and Y.~Zhang,
\newblock A review of deep-learning-based medical image segmentation methods,
\newblock Sustainability {\bf 13}, 1224 (2021).

\bibitem{sahiner2019deep}
B.~Sahiner, A.~Pezeshk, L.~M. Hadjiiski, X.~Wang, K.~Drukker, K.~H. Cha, R.~M. Summers, and M.~L. Giger,
\newblock Deep learning in medical imaging and radiation therapy,
\newblock Medical physics {\bf 46}, e1--e36 (2019).

\bibitem{ronneberger2015unet}
O.~Ronneberger, P.~Fischer, and T.~Brox,
\newblock U-Net: Convolutional Networks for Biomedical Image Segmentation,
\newblock arXiv preprint arXiv:1505.04597  (2015).

\bibitem{li2021gt}
Y.~Li, S.~Wang, J.~Wang, G.~Zeng, W.~Liu, Q.~Zhang, Q.~Jin, and Y.~Wang,
\newblock Gt u-net: A u-net like group transformer network for tooth root segmentation,
\newblock in {\em Machine Learning in Medical Imaging: 12th International Workshop, MLMI 2021, Held in Conjunction with MICCAI 2021, Strasbourg, France, September 27, 2021, Proceedings 12}, pages 386--395, Springer, 2021.

\bibitem{li2023lvit}
Z.~Li, Y.~Li, Q.~Li, P.~Wang, D.~Guo, L.~Lu, D.~Jin, Y.~Zhang, and Q.~Hong,
\newblock Lvit: language meets vision transformer in medical image segmentation,
\newblock IEEE transactions on medical imaging  (2023).

\bibitem{zhou2019unet++}
Z.~Zhou, M.~M.~R. Siddiquee, N.~Tajbakhsh, and J.~Liang,
\newblock Unet++: Redesigning skip connections to exploit multiscale features in image segmentation,
\newblock IEEE transactions on medical imaging {\bf 39}, 1856--1867 (2019).

\bibitem{wang2024amsc}
Y.~Wang, R.~Dan, S.~Luo, L.~Sun, Q.~Wu, Y.~Li, X.~Chen, K.~Yan, X.~Ye, and D.~Yu,
\newblock AMSC-Net: Anatomy and multi-label semantic consistency network for semi-supervised fluid segmentation in retinal OCT,
\newblock Expert Systems with Applications , 123496 (2024).

\bibitem{siddique2021u}
N.~Siddique, S.~Paheding, C.~P. Elkin, and V.~Devabhaktuni,
\newblock U-net and its variants for medical image segmentation: A review of theory and applications,
\newblock Ieee Access {\bf 9}, 82031--82057 (2021).

\bibitem{chen2021TransUNet}
J.~Chen, Y.~Lu, Q.~Yu, X.~Luo, E.~Adeli, Y.~Wang, L.~Lu, A.~L. Yuille, and Y.~Zhou,
\newblock TransUNet: Transformers Make Strong Encoders for Medical Image Segmentation,
\newblock arXiv preprint arXiv:2102.04306  (2021).

\bibitem{cao2021swin}
H.~Cao, Y.~Wang, J.~Chen, D.~Jiang, X.~Zhang, Q.~Tian, and M.~Wang,
\newblock Swin-Unet: Unet-like Pure Transformer for Medical Image Segmentation,
\newblock arXiv preprint arXiv:2105.05537  (2021).

\bibitem{isensee2020nnUNet}
F.~Isensee, P.~F. Jaeger, S.~A.~A. Kohl, J.~Petersen, and K.~H. Maier-Hein,
\newblock nnU-Net: a self-configuring method for deep learning-based biomedical image segmentation,
\newblock Nature Methods {\bf 17}, 203--211 (2020).

\bibitem{kirillov2023segment}
A.~Kirillov et~al.,
\newblock Segment anything,
\newblock arXiv preprint arXiv:2304.02643  (2023).

\bibitem{zhang2023faster}
C.~Zhang, D.~Han, Y.~Qiao, J.~U. Kim, S.-H. Bae, S.~Lee, and C.~S. Hong,
\newblock Faster Segment Anything: Towards Lightweight SAM for Mobile Applications,
\newblock arXiv preprint arXiv:2306.14289  (2023).

\bibitem{mazurowski2023segment}
M.~A. Mazurowski, H.~Dong, H.~Gu, J.~Yang, N.~Konz, and Y.~Zhang,
\newblock Segment anything model for medical image analysis: an experimental study,
\newblock Medical Image Analysis {\bf 89}, 102918 (2023).

\bibitem{ma2023segment}
J.~Ma, Y.~He, F.~Li, L.~Han, C.~You, and B.~Wang,
\newblock Segment Anything in Medical Images,
\newblock arXiv preprint arXiv:2304.12306  (2023).

\bibitem{shaharabany2023autosam}
T.~Shaharabany, A.~Dahan, R.~Giryes, and L.~Wolf,
\newblock AutoSAM: Adapting SAM to Medical Images by Overloading the Prompt Encoder,
\newblock arXiv preprint arXiv:2306.06370  (2023).

\bibitem{li2011level}
C.~Li, R.~Huang, Z.~Ding, J.~C. Gatenby, D.~N. Metaxas, and J.~C. Gore,
\newblock A level set method for image segmentation in the presence of intensity inhomogeneities with application to MRI,
\newblock IEEE transactions on image processing {\bf 20}, 2007--2016 (2011).

\bibitem{wang2020deep}
Y.~Wang, X.~Wei, F.~Liu, J.~Chen, Y.~Zhou, W.~Shen, E.~K. Fishman, and A.~L. Yuille,
\newblock Deep distance transform for tubular structure segmentation in ct scans,
\newblock in {\em Proceedings of the IEEE/CVF Conference on Computer Vision and Pattern Recognition}, pages 3833--3842, 2020.

\bibitem{luo2021semi}
X.~Luo, J.~Chen, T.~Song, and G.~Wang,
\newblock Semi-supervised medical image segmentation through dual-task consistency,
\newblock in {\em Proceedings of the AAAI conference on artificial intelligence}, volume~35, pages 8801--8809, 2021.

\bibitem{dosovitskiy2020image}
A.~Dosovitskiy et~al.,
\newblock An image is worth 16x16 words: Transformers for image recognition at scale,
\newblock arXiv preprint arXiv:2010.11929  (2020).

\bibitem{wu2022tinyvit}
K.~Wu, J.~Zhang, H.~Peng, M.~Liu, B.~Xiao, J.~Fu, and L.~Yuan,
\newblock Tinyvit: Fast pretraining distillation for small vision transformers,
\newblock in {\em European Conference on Computer Vision}, pages 68--85, Springer, 2022.

\bibitem{chen2020learning}
X.~Chen, X.~Luo, Y.~Zhao, S.~Zhang, G.~Wang, and Y.~Zheng,
\newblock Learning Euler's elastica model for medical image segmentation,
\newblock arXiv preprint arXiv:2011.00526  (2020).

\bibitem{zhuang2019evaluation}
X.~Zhuang et~al.,
\newblock Evaluation of algorithms for multi-modality whole heart segmentation: an open-access grand challenge,
\newblock Medical image analysis {\bf 58}, 101537 (2019).

\bibitem{crimi2019brainlesion}
A.~Crimi, S.~Bakas, H.~Kuijf, F.~Keyvan, M.~Reyes, and T.~van Walsum,
\newblock {\em Brainlesion: Glioma, Multiple Sclerosis, Stroke and Traumatic Brain Injuries: 4th International Workshop, BrainLes 2018, Held in Conjunction with MICCAI 2018, Granada, Spain, September 16, 2018, Revised Selected Papers, Part II}, volume 11384,
\newblock Springer, 2019.

\bibitem{jaeger2013automatic}
S.~Jaeger et~al.,
\newblock Automatic tuberculosis screening using chest radiographs,
\newblock IEEE transactions on medical imaging {\bf 33}, 233--245 (2013).

\bibitem{oktay2022attention}
O.~Oktay et~al.,
\newblock Attention U-Net: Learning Where to Look for the Pancreas,
\newblock in {\em Medical Imaging with Deep Learning}, 2022.

\bibitem{li2021agmb}
Y.~Li et~al.,
\newblock Agmb-transformer: Anatomy-guided multi-branch transformer network for automated evaluation of root canal therapy,
\newblock IEEE Journal of Biomedical and Health Informatics {\bf 26}, 1684--1695 (2021).

\end{thebibliography}
\bibliographystyle{./medphy.bst}

\end{document}